\documentclass[preprint,12pt]{elsarticle}

\usepackage{amsmath}%
\journal{Neural Networks}

\begin{document}

\begin{frontmatter}


\title{Spiking Two-Stream Methods with Unsupervised STDP-based Learning for Action Recognition}
%\tnotetext[label1]{}
\author[label1]{Mireille El-Assal\corref{cor1}}
\ead{mireille.elassal2@univ-lille.fr}
\author[label1]{Pierre Tirilly}
\ead{pierre.tirilly@univ-lille.fr}
\author[label1]{Ioan Marius Bilasco}
\ead{marius.bilasco@univ-lille.fr}
\affiliation[label1]{organization={Univ. Lille, CNRS, Centrale Lille, MR 9189 -- CRIStAL -- Centre de Recherche en Informatique, Signal et Automatique de Lille},
            city={Lille},
            postcode={F-59000},
            country={France}}

\cortext[cor1]{Corresponding author.}
\begin{abstract}
Video analysis is a computer vision task that is useful for many applications like surveillance, human-machine interaction, and autonomous vehicles. Deep Convolutional Neural Networks (CNNs) are currently the state-of-the-art methods for video analysis. However they have high computational costs, and need a large amount of labeled data for training. In this paper, we use Convolutional Spiking Neural Networks (CSNNs) trained with the unsupervised Spike Timing-Dependent Plasticity (STDP) learning rule for action classification. These networks represent the information using asynchronous low-energy spikes. This allows the network to be more energy efficient and neuromorphic hardware-friendly. However, the behaviour of CSNNs is not studied enough with spatio-temporal computer vision models. Therefore, we explore transposing two-stream neural networks into the spiking domain. Implementing this model with unsupervised STDP-based CSNNs allows us to further study the performance of these networks with video analysis. In this work, we show that two-stream CSNNs can successfully extract spatio-temporal information from videos despite using limited training data, and that the spiking spatial and temporal streams are complementary. We also show that using a spatio-temporal stream within a spiking STDP-based two-stream architecture leads to information redundancy and does not improve the performance.
\end{abstract}

% \begin{highlights}
%  \item We present a two-stream spiking convolutional STDP-based model that is explored with five different types of temporal streams;
%  \item we evaluate and compare the performance of two-stream spiking convolutional models with our different temporal streams on the KTH, Weizmann, IXMAS and UCF sports, action recognition datasets;
%  \item we conclude that spiking two-stream methods can successfully extract spatio-temporal features from action recognition videos; 
%  \item we conclude that spiking spatial and temporal streams are complementary with all of the chosen temporal stream configurations;
%  \item we conclude that spatio-temporal streams (like 3D CSNNs) lead to information redundancy and does not give effective results with STDP-based two-stream spiking models;
%  \item and we conclude that these models are sensitive to significant spatial noise which has a negative impact on the performance.


\end{frontmatter}


\section{Introduction}
\label{sec:intro}
Large amounts of new visual data are made publicly available to the world on a daily basis. This has encouraged the development of algorithms capable of analyzing this data, like Artificial Neural Networks (ANNs). Convolutional Neural Networks (CNNs) have shown impressive performance with image analysis tasks like image classification or segmentation~\citep{ImgClassNIPS, imgoverfeat}. This, in turn, has encouraged their use with video analysis~\citep{LSvidclassCNN}, which requires the network to take the temporal relationship between the video frames into account. 
There are different approaches for information processing over the temporal dimension using CNNs, like fusion techniques~\citep{LSvidclassCNN}, and using two stream methods~\citep{LSvidclassCNN, two-streamCNN}.
However, traditional deep neural networks have some drawbacks like their considerable computational costs~\citep{TAVANAEI201947}, so implementing them on devices using limited energy can be a problem. Moreover, they are trained with supervised learning algorithms like gradient descent back-propagation, which require large amounts of labeled data that call for costly human intervention. These issues of traditional CNNs can be avoided by using Convolutional Spiking Neural Networks (CSNNs). 

CSNNs process visual data using low-energy action potentials referred to as spikes, instead of regular values. These models have potential benefits, like fast information processing when implemented on neuromorphic hardware, and energy efficiency~\citep{TAVANAEI201947},~\citep{multLyrSNNWithTargetTmStampTrshAdpt}. SNN training methods like ANN-to-SNN conversion~\citep{ann-to-snn-vid} and spatio-temporal back-propagation~\citep{STBPHighPerformenceSNN} do not entirely overcome ANN bottlenecks like the need for large amounts of labeled data for training, and the fact that they are more difficult to implement on neuromorphic hardware than Hebbian methods. On the other hand, training unsupervised SNNs with Hebbian learning rules like Spike Timing-Dependent Plasticity (STDP) allows computations to be done locally, which then makes their implementation possible on a wide variety of low power devices. Although these STDP-based models do not yet yield competitive performances with ANNs~\citep{PFalezVisualFeatureLearning}, understanding their performance with spatio-temporal architectures can help bridge this performance gap. 

In this work, we learn spatio-temporal features without supervision from human action videos by transferring into the spiking domain an effective architecture used in CNN-based computer vision: the two-stream architecture. We use CSNNs trained with the unsupervised STDP learning rule~\citep{ImprSNNTrain} for the streams considered.
We evaluate the performance of this model on the KTH~\citep{RecogHumanActions}, Weizmann~\citep{ActionsAsSpaceTimeShapes}, IXMAS~\citep{weinland:inria-ixmas}, and UCF sports~\citep{ufcsport} datasets.

This work introduces a new building block towards developing unsupervised STDP-based spiking models that can learn spatio-temporal features. The main contributions of this paper are summarized as follows: 
\begin{itemize} %that allows learning spatio-temporal patterns with STDP in an unsupervised manner
  \item we present a two-stream spiking convolutional model, and we investigate motion modeling with five different types of temporal streams in Section~\ref{section:two-stream};
  \item we evaluate and compare the performance of two-stream spiking convolutional models with our different temporal streams on the KTH, Weizmann,  UCF sports and IXMAS action recognition datasets in Section~\ref{section:evaluation}.
\end{itemize}

\section{Related Work}
\label{section:literature}
\subsection{Video analysis with traditional neural networks}
\label{subsection:mmodelingANN}
Several deep approaches are being used for video analysis, like deep belief networks~\citep{RBM_HAR}, and Recurrent Neural Networks (RNNs)~\citep{RNNHumanMotionPrediction, Zhang_2017_ICCV, RNN_HAR_JAOUEDI2020447}, which show good performance for a variety of sequence analysis tasks. Long-Short Term Memory (LSTM) cells are often combined with RNNs, where the LSTM serves as the memory units for the gradient descent~\citep{RNN_HAR_Singh_2017, RNN_LSTM_HAR_hammerla2016, RNN_LSTM_HAR_7743581}. However, these models are most effective with short sequences, while convolutional neural networks are more suitable for learning long-term repetitive sequences~\citep{HAR_sensor_WANG20193}. 3D CNNs can naturally extract spatio-temporal features from human action recognition videos with their third dimension dedicated to time~\citep{can3Dret2D, slowfast_ICCV, X3D, LearningSPTFeatures}. However, these networks need large amounts of training data in order to achieve good performance.

Multi-stream networks have shown good performance with video analysis, even when using limited training data~\citep{LSvidclassCNN},~\citep{two-streamCNN}.  Some argue that these networks have more benefits than other methods that extend CNNs for video classification, like reduced run-time~\citep{LSvidclassCNN}. These networks extract different features using two different pathways, usually a spatial one and a temporal one. The spatial stream extracts spatial information from a single video frame~\citep{two-streamCNN}, while the temporal stream  can have many different configuration options. Some authors use RNNs, LSTMs and 3D CNNs as streams in two-stream network architectures~\citep{Wang2017ModelingTD, liu2017twostream, feichtenhofer2016convolutional}; they show a benefit in using spatio-temporal streams that already extract spatio-temporal features. Some work also uses pose or skeleton information, like in~\citep{liu2017twostream}, where the authors find that two-stream methods that are based on 3D CNNs can be improved by adding a stream for pose estimation. In the case of action recognition, actions can manifest significant variations in appearance and motion across different instances of a video. By modeling spatial and temporal aspects separately, two-stream networks can better handle these variations, which leads to more robustness. %These networks can also benefit from pre-training on large-scale image datasets (e.g., ImageNet) for the spatial stream and on large-scale video datasets (e.g., Kinetics) for the temporal stream}. 
In~\citep{Ye_2015}, the authors implement two-stream CNNs for action recognition. They compare the performance of features that are extracted from different layers of spatial and temporal CNNs, and they use an SVM for classification. They conclude that fusing features from a spatial stream with those from a temporal stream is beneficial. However, fusing a network with strong performance with another one of weak performance is not helpful in improving the overall fusion performance.

However, all of the neural networks mentioned in this section use regular analog values for information processing, and not spikes. These models still require a large number of labeled samples for training, and are generally not natively energy efficient. In this work, we address the need for cost efficient and unsupervised motion modeling, by using SNNs that make this possible.

\subsection{Video analysis with spiking neural networks}
\label{subsection:mmodelingSNN}

Spiking neural networks are third generation neural networks that have received great interest recently in the domain of video analysis, especially for their ability to deal with event data from dynamic vision sensors~\citep{EventBasedVision, DSNNMotionSeg}. However, the use of these networks should not be limited to spiking cameras, because there are large amounts of frame-based video data in need of analysis. Spiking networks have been explored with RNNs~\citep{newSCRNN}, as well as 3D CNNs~\citep{elassal:hal-03679597IJCNN}. In~\citep{elassal:hal-03679597IJCNN}, the network is trained using unsupervised STDP learning, and can extract spatio-temporal features naturally from videos. There is very little work in the literature in which spiking two-stream methods are used. In~\citep{two-strRSNN}, the authors introduce a deep two-stream SNN based on a spiking ResNet50 architecture and a recurrent spiking neural network (RSNN) fusion module. They use ANN-to-SNN conversion with their own hybrid conversion method. However, ANN-to-SNN conversion still requires the training to be done with a non-spiking ANN, thus reducing the energy efficiency benefits. In~\citep{Zhu_2021_ICCV}, the authors extract two different types of data from neuromorphic sensors, in order to reconstruct 2D images from spiking images.
They extract a motion path and a texture path from the same event stream and fuse them together. Then, they use a NeuSpike-Net, that has two encoders and a decoder. However, this work does not address spiking two-stream methods, because their network is not a spiking model.

To the best of our knowledge, two-stream methods have not yet been explored with unsupervised STDP-based learning in the spiking domain, and this is the first work that explores STDP-based two-stream methods for unsupervised convolutional spiking neural networks.

In our work, we are inspired by the two-stream methods discussed in Section~\ref{subsection:mmodelingANN}, and find it necessary to use these methods with unsupervised STDP-based convolutional SNN in order to decrease their computational and labeling costs. Therefore, we introduce two-stream convolutional spiking  neural networks trained with the unsupervised STDP learning rule. We test multiple possible temporal streams, and we use both 2D and 3D architectures for temporal and spatio-temporal feature extraction.

\section{Background \& Network Architecture}
\label{section:networkarch}

This section introduces the needed background information and the mechanisms chosen to achieve unsupervised spatio-temporal feature extraction with STDP.

\subsection{A video sample}
\label{subsection:video}
A video is a sequence of frames that is represented as a 4D tensor of size $l_{w} \times l_{h} \times l_{c} \times l_{td}$ where $l_{w}$ and $l_{h}$ are the width and height of a video frame, $l_{c}$ is the number of channels (e.g. three for RGB frames), and $l_{td}$ is the temporal depth, which is the number of frames in the video sample.

\begin{figure*}
\centerline{\includegraphics[scale=0.31]{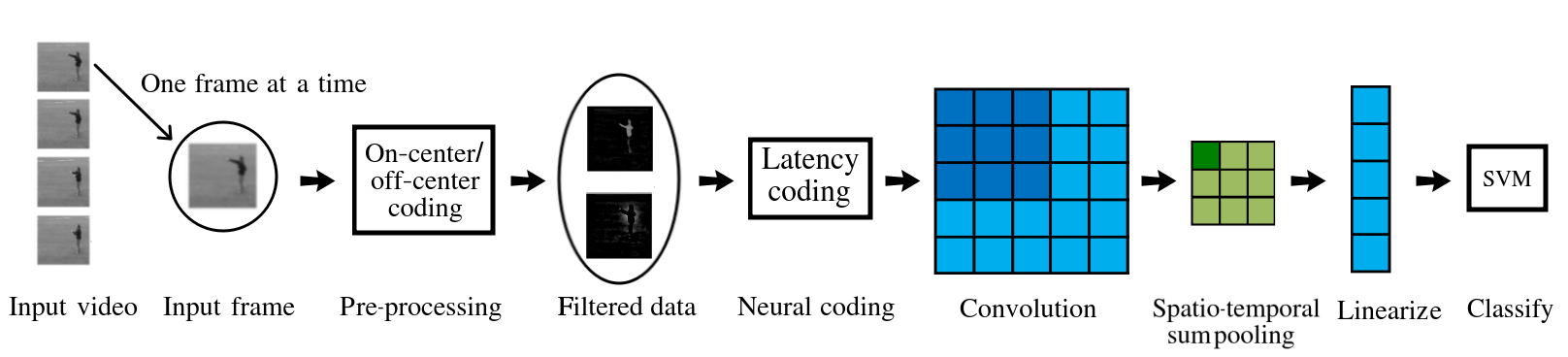}}
\caption{Single layer convolutional spiking neural network used in this work.}
\label{figNetwork}
\end{figure*}

\subsection{Baseline architecture}
\label{subsection:baselinearch}
The single layer general architecture used in this work is shown in Figure~\ref{figNetwork}. A video sample is pre-processed one frame at a time, and then the information is converted into spikes as explained in Section~\ref{subsubsection:onofflatency}. These spikes are then processed by the CSNN, and they produce output spiking feature maps that are transformed back from spikes into regular values, before using sum pooling to decrease their size. They are then introduced into a Support Vector Machine (SVM). 
The SVM is needed to classify the extracted features, because feature learning is unsupervised in this work. The final classification method could be any supervised method; we chose to use an SVM because it is a standard approach that is effective with default hyper-parameters. 

\subsubsection{On-center/off-center filtering and neural coding}
\label{subsubsection:onofflatency}
In this work, input video frames are pre-processed using an on-center/off-center filter~\citep{OnOffCS}. This filter is a difference-of-Gaussian (DoG) filter that is used to pre-process the data by simulating on-center/off-center cells and extracting edges. This filter reproduces the behaviour of the retina, and is represented by Equation~\ref{eq:onoff} from~\citep{ImprSNNTrain}:

\begin{equation} \label{eq:onoff}
DoG_{s, \sigma_{1},\sigma_{2}}(x,y) = I(x,y)\ast(G_{s,\sigma_{1}} - G_{s,\sigma_{2}})
\end{equation}

where $I$ is the input image, $\ast$ is the convolution operator and $G_{s,\sigma}$ is a zero-mean normalized Gaussian kernel of size $s \times s$ and scale $\sigma$. Then the positive and negative values generated from this filtering process are separated into two channels, where the positive one represents the on cells, and the negative one represents the off cells as shown in Equations~\ref{eq:on}~and~\ref{eq:off}~from~\citep{ImprSNNTrain}:
\begin{equation} \label{eq:on}
x_{on} = max(0, DoG(x, y))
\end{equation}
\begin{equation} \label{eq:off}
x_{off} = max(0, -DoG(x, y))
\end{equation}

The resulting pre-processed frames are then transformed into spikes using a temporal neural coding called latency coding~\citep{guo_neural_2021}, as shown in Equation~\ref{eq:lc}~\citep{ImprSNNTrain}.%\citep{THORPE2001715}.
\begin{equation} \label{eq:lc}
f_{\mathrm{in}}(x) = (1.0 - x)\times t_{\mathrm{exposition}}
\end{equation}

where $x \in [0, 1]$ is the input pixel value, and $t_{\mathrm{exposition}}$ represents the presentation duration for one sample, which is set to $1$ in this work. 

\subsubsection{Neuron model}
The neural networks proposed in this work consist of a feed-forward layer of integrate-and-fire (IF) neurons~\citep{IFneuron}. The IF neuron model is the simplest and the most used to learn visual features. Its simplicity makes it easy to implement with neuromorphic hardware. It has a membrane potential $v(t)$, a membrane threshold $v_{\text{th}}(t)$, and a membrane capacitance $C_{m}$, which we set to $1$ in this paper. 
Input spikes increase the neuron membrane potential until it reaches the threshold potential, thus prompting the neuron to fire an output spike. Then, the membrane potential of the neuron goes back to the resting potential $v_{r}$, which we set to $0$ volts. The behaviour of this neuron is characterised by Equations~\ref{eq:vtIF}~and~\ref{eq:fskernel}: %~\citep{ImprSNNTrain}:
\begin{equation} \label{eq:vtIF}
\begin{array}{c}
\displaystyle v(t) = \sum_{i \in \mathcal{E}} w_{i} f_{s}(t-t_{i})\\ v(t) \leftarrow v_{r} \mbox{ when }  v(t) \geq v_{\text{th}}(t) 
\end{array}
\end{equation}
\begin{equation}\label{eq:fskernel}
f_{s}(x) = 
\begin{cases}
    1, & \mbox{if } x \geq 0 \\ 
    0, & \mbox{otherwise}
\end{cases}
\end{equation}
where $\mathcal{E}$ represents the set of incoming spikes, $w_{i}$ is the weight of the synapse carrying the \(i\)-th spike, $t_{i}$ represents the timestamp of the \(i\)-th spike, and $f_{s}$ is the kernel of spikes. 

\subsubsection{STDP learning rule}
Random patches of the spiking sample are selected to train the CSNN, similarly to some work with STDP in the literature~\citep{SCNN-STDP-img, ImprSNNTrain}. We use the unsupervised biological STDP learning rule for training~\citep{SchumanSTDP}, which performs better than both multiplicative and additive STDP, by adding non-linearity, as explained in~\citep{ImprSNNTrain}. It allows the learning of more complex features. This learning rule is characterised by Equation~\ref{eq:lr}~\citep{ImprSNNTrain}:

\begin{equation} \label{eq:lr}
\Delta_{w} = 
\begin{cases}
    \eta_{w} e^{-\frac{t_{ \text{post}} - t_{\text{pre}}}{\tau_{\text{STDP}}}}, & \mbox{if } t_{\text{pre}} \leq t_{\text{post}} \\ 
    -\eta_{w} e^{-\frac{t_{\text{pre}} - t_{\text{post}}}{\tau_{\text{STDP}}}}, &  \text{otherwise}
\end{cases}
\end{equation}

where $t_{\text{pre}}$ and $t_{\text{post}}$ are the timestamps for input and output spikes respectively, $\eta_{w}$ is the learning rate and $\tau_{\text{STDP}}$ is the time constant responsible for the STDP learning window. 

\subsubsection{Threshold adaptation}
\label{subsubsection:threshadapt}
To prevent several neurons from learning similar patterns, this neural network uses the winner-takes-all (WTA) inhibition method~\citep{wta} during training. However, with this inhibition method, some neurons can overpower other neurons by having a tendency to fire more spikes. This can cause the network to become stuck in a state where the same few active neurons are firing all the time, leaving the other neurons quiet. In order to avoid this and ensure the stability of the network, a homeostasis mechanism is needed: we use the threshold adaptation method introduced in~\citep{AdaptiveTH}. We also use the target timestamp threshold adaptation method from~\citep{multLyrSNNWithTargetTmStampTrshAdpt}. This method prompts the neurons to fire at a pre-defined target timestamp $\hat{t}$ that allows learning specific patterns while maintaining the homeostasis of the network. The thresholds of all neurons (winners and losers), are adapted each time a neuron fires or receives an inhibitory spike, so that their firing time converges towards this target timestamp~\citep{ImprSNNTrain}. The thresholds are updated according to Equations~\ref{eq:deltaTH1},~\ref{eq:deltaTH2},~and~\ref{eq:vth}:
\begin{equation} \label{eq:deltaTH1}
\Delta_{\text{th}}^1 = -\eta_{\text{th}}(t - \hat{t})
\end{equation}

\begin{equation} \label{eq:deltaTH2}
\Delta_{\text{th}}^2 = 
\begin{cases}
    \eta_{\text{th}}, & \mbox{if } \mbox{$t_{i} = \text{min}\{t_{0},...,t_{N}\}$} \\ 
    -\frac{\eta_{\text{th}}}{l_{d}(n)}, & \mbox{otherwise}
\end{cases}
\end{equation}

\begin{equation} \label{eq:vth}
v_{\text{th}}(t) = \text{max}(\text{th}_{\text{min}}, v_{\text{th}}(t-1)+ \Delta_{\text{th}}^1 + \Delta_{\text{th}}^2) 
\end{equation}

where $t$ is the timestamp at which the neuron fires, $\eta_{\text{th}}$ is the threshold learning rate, $l_{d}$ is the number of neurons that are in competition in the layer, $t_{i}$ is the firing timestamp of neuron $i$, $\text{min}\{t_{0},...,t_{N}\}$ is the minimum timestamp which corresponds to the neuron that fired first, and $\text{th}_{\text{min}}$ is the minimum possible threshold value~\citep{ImprSNNTrain}.

\section{Two-stream methods}
\label{section:two-stream}
\begin{figure}
\centerline{\includegraphics[scale=0.23]{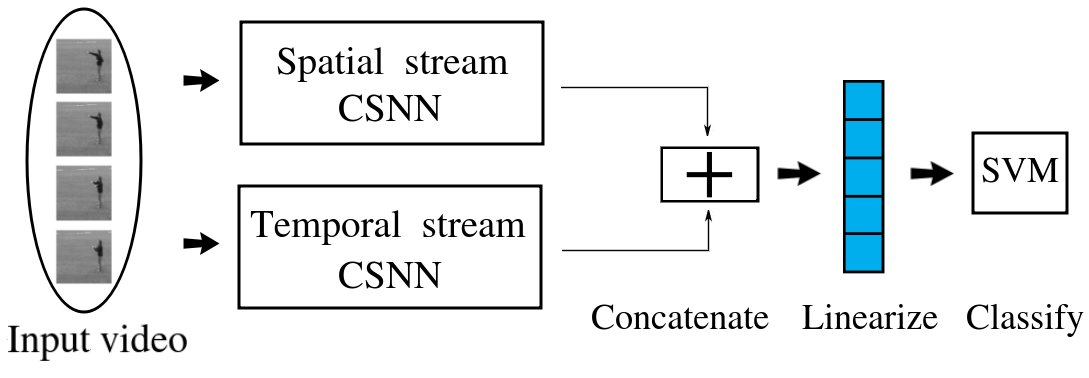}}
\caption{General two-stream architecture used in this paper.}
\label{figTS}
\end{figure}

Two-stream networks have two different streams used for extracting different types of features from the same video sample, as shown in Figure~\ref{figTS}. The appearance information is extracted by the spatial stream, and the motion information is extracted by the temporal stream. 
Then the information extracted from both streams is concatenated and sent into the classifier.

\subsection{The spatial stream}
\begin{figure*}
\centerline{\includegraphics[scale=0.23]{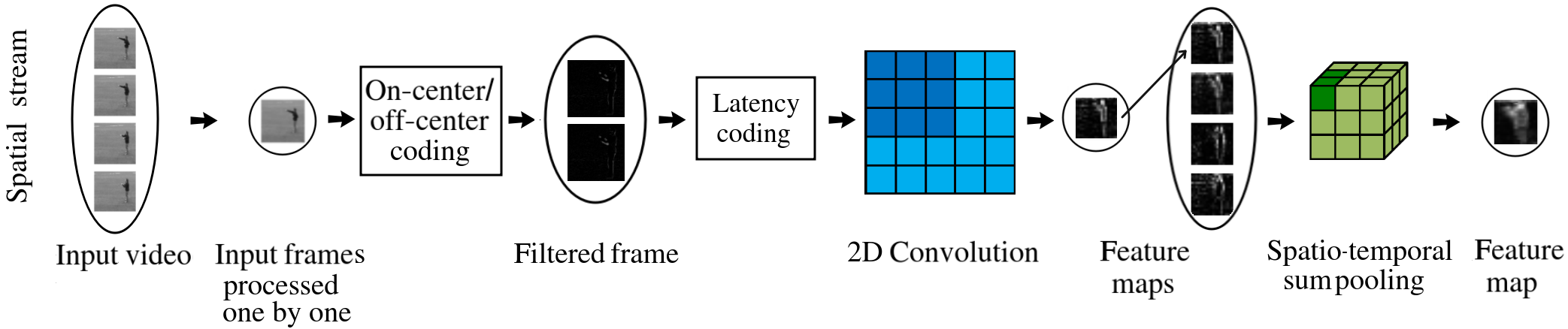}}
\caption{Architecture of the spatial stream.}
\label{figTSsp}
\end{figure*}

The spatial stream used in this work is a 2D CSNN. This stream is responsible for extracting appearance information from the spiking data. This stream produces a spiking feature map as an output for each frame in the video. Then these feature maps are transformed back into regular non-spiking values and undergo temporal sum-pooling to become a single feature map that represents the entire clip, as shown in Figure~\ref{figTSsp}. This single feature map is then concatenated with the output of the temporal stream.

\subsection{The temporal stream}
In this work, we explore five different streams that allow us to extract temporal data from videos.

\begin{figure*}
\centerline{\includegraphics[scale=0.23]{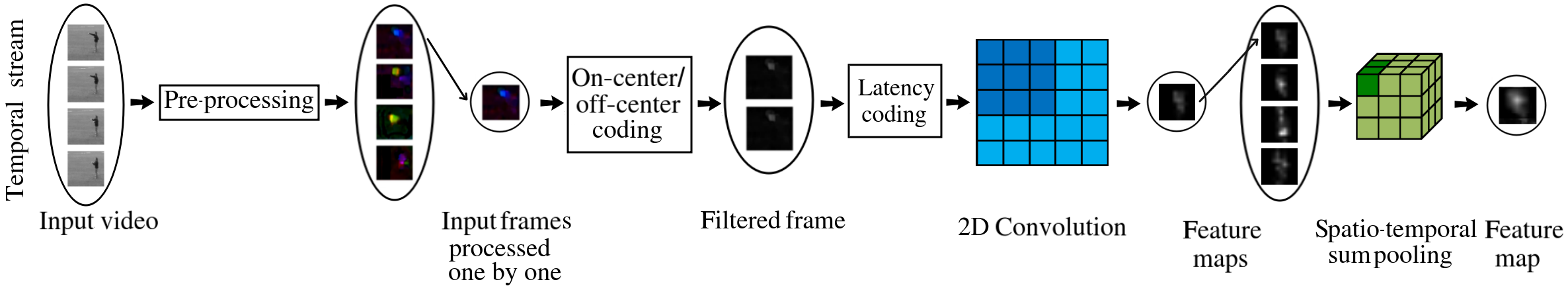}}
\caption{Architecture of the temporal streams based on 2D convolutions.}
\label{figTStmp}
\end{figure*}

\subsubsection{Early fusion}
\begin{figure}
\centerline{\includegraphics[scale=0.45]{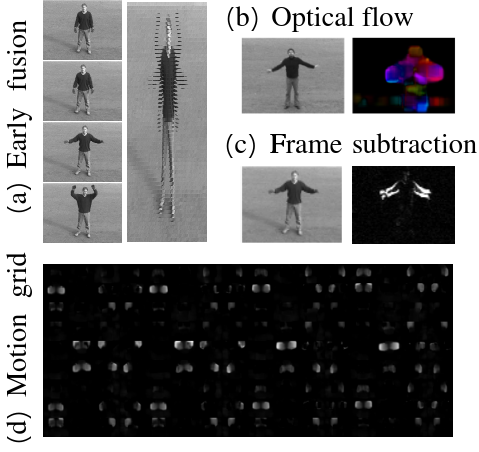}}
\caption{ (a) Early fusion. (b) Frame subtraction. (c) Optical flow. (d) Motion grid.}
\label{fig:preproc}
\end{figure}

Fusion is one way of aggregating the temporal information in a sequence of frames. In deep neural networks, there are multiple types of fusion~\citep{LSvidclassCNN}. We are interested in fusion as a pre-processing, so we use early fusion. However, the early fusion technique mentioned in~\citep{LSvidclassCNN} includes using a convolution operation in the temporal dimension. In this paper, we want to model the motion present in a video sample as a static representation of motion before any processing by the CSNN, so we implement an early fusion method from~\citep{elassal:hal-03263914}. This allows us to use early fusion as a pre-processing method without changing the architecture of the 2D CSNN. This early fusion method is achieved by fusing multiple frames together row by row, thus forming one large frame, as shown in Figure~\ref{fig:preproc}(a). This is represented by Equation~\ref{eq:ef}:
\begin{equation} \label{eq:ef}
\begin{array}{l}
I_{o}(r,y) = I^n(x,y) 
\text{ with } r=x \times l_{td}+n, \\i \in [0,l_{td}-1], x \in [0,l_{h}-1], y \in [0,l_{w}-1]
\end{array}
\end{equation}
where $l_{w}$ and $l_{h}$ are the width and height of the input frame, $l_{td}$ is the number of frames, $I^n$ is the input frame of index $n$, $I_{o}$ is the output frame of width $l_{w}$ and height $l_{h} \times l_{td}$, $r$ is the row index, and $x$ and $y$ are the pixel coordinates in the vertical and horizontal dimensions, respectively.

\subsubsection{Optical flow}
Optical flow is the most commonly used form of temporal stream information with two-stream methods~\citep{two-streamCNN,TS3D2,tmpTS_HAR}. A pre-processing method based on Farneback's dense optical flow~\citep{twoFrameMotionEstimation} is applied to the temporal stream in our work. We separate optical flow vectors into orientation and magnitude values. The orientation data is periodic, therefore, it is difficult to apply latency coding to it. Thus, the information is displayed in the HSV color space, which is then converted into RGB color space, as shown in Figure~\ref{fig:preproc}(b) before latency coding is applied~\citep{elassal:hal-03263914}.

\subsubsection{Frame subtraction}
Another method to extract motion information is to simply subtract each two consecutive frames. This provides positive and negative values depending on the pixel values. Since temporal coding is only defined for positive input values, we take the absolute value of the negative values, as shown in Equation~\ref{eq:fs}:  
\begin{equation} \label{eq:fs}
    I^n_{o}(x,y) = |I^n(x,y) - I^{n+1}(x,y)|
\end{equation}
where $I^n$ is the input frame of index $n \in [0, l_{td}-2]$, $I^n_{o}$ is the output frame resulting from frame subtraction, and $(x,y)$ are the pixel coordinates. Moreover, subtracting consecutive frames removes the static spatial information like background pixels, and parts of the subject that do not move, as shown in Figure~\ref{fig:preproc}(c).

\subsubsection{Motion grid} This method groups the movement information of several optical flow frames into a composite grid, and is the pre-processing method that gave the best performance with STDP-based SNNs in~\citep{elassal:hal-03263914}. Each optical flow frame is separated into four separate sub-frames representing the displacement in four different directions one after the other, as shown in Figure~\ref{fig:preproc}(d). The displacements of motion used in this method are represented in Equations~\ref{eq:ml},~\ref{eq:mr},~\ref{eq:mu},~and~\ref{eq:md}:

\begin{equation} \label{eq:ml}
M_{l}(x,y) = \frac{|\text{OF}_{x}(x,y)| - \text{OF}_{x}(x,y)}{2}
\end{equation}
\begin{equation}\label{eq:mr}
M_{r}(x,y) = \frac{|\text{OF}_{x}(x,y)| + \text{OF}_{x}(x,y)}{2}
\end{equation}
\begin{equation}\label{eq:mu}
M_{u}(x,y) = \frac{|\text{OF}_{y}(x,y)| - \text{OF}_{y}(x,y)}{2}
\end{equation}
\begin{equation}\label{eq:md}
M_{d}(x,y) = \frac{|\text{OF}_{y}(x,y)| + \text{OF}_{y}(x,y)}{2}
\end{equation}

where $\text{OF}_{x}$ and $\text{OF}_{y}$ represent the horizontal and vertical components of the extracted optical flow, $M_{u}(x,y)$ is the upwards displacement at pixel $(x, y)$, $M_{d}(x,y)$ the downwards displacement, $M_{l}(x,y)$ is the displacement to the left, and $M_{r}(x,y)$ is the displacement to the right.

\subsubsection{3D convolution}
\begin{figure*}
\centerline{\includegraphics[scale=0.28]{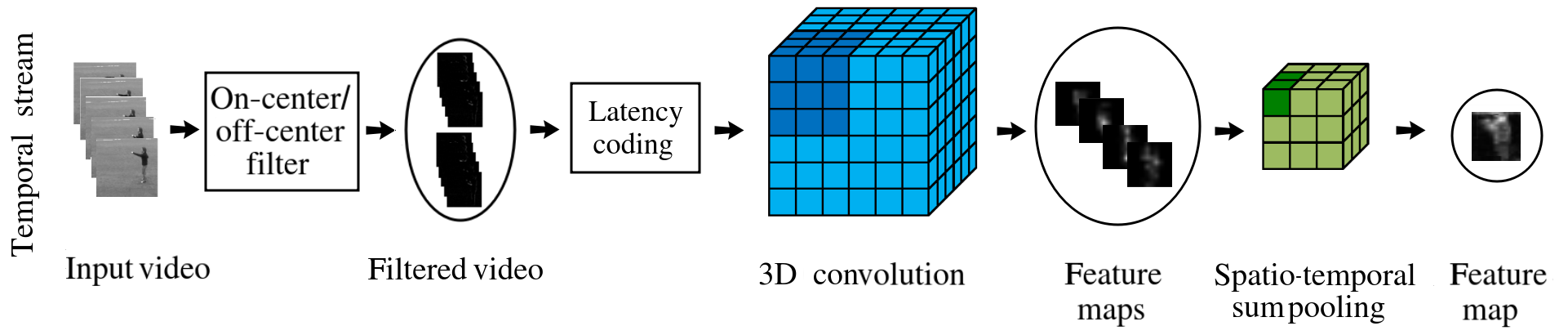}}
\caption{Architecture of the temporal stream based on 3D convolutions.}
\label{fig6}
\end{figure*}

With two-stream methods, the temporal stream was originally intended to only extract temporal features, while the appearance information is extracted by the spatial stream. Yet, authors claim that it is more effective to fuse a spatial stream with a spatio-temporal stream that uses 3D CNNs~\citep{liu2017twostream}. In this work, we replace the temporal stream by the 3D convolutional spiking neural network from~\citep{elassal:hal-03679597IJCNN} to extract spatio-temporal information. This is to see if the same amelioration in performance is attainable in the spiking domain. Figure~\ref{fig6} shows the temporal stream used in this case.

\begin{figure}
\centerline{\includegraphics[scale=0.2]{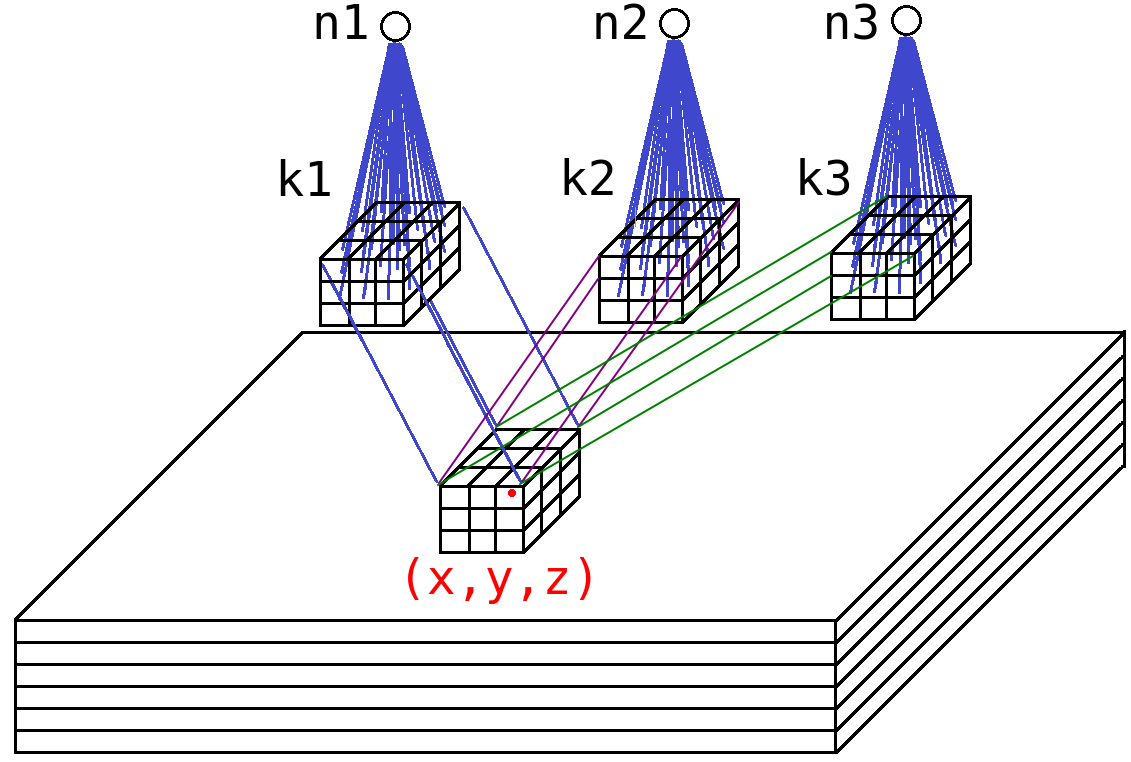}}
\caption{3D convolution: the neighbourhood of a neuron at \((x, y, z)\) is a volume, and n1, n2, and n3 are competing neurons at the same location~\citep{elassal:hal-03679597IJCNN}.}
\label{fig:3DRS}
\end{figure}

A 3D spiking convolutional layer has $f_{k}$ trainable kernels, with sizes $f_{w} \times f_{h} \times f_{\text{td}}$, where $f_{w}$ and $ f_{h}$ represent the width and height of the kernel respectively and $f_{\text{td}}$ is the temporal size of the kernel, as shown in Figure~\ref{fig:3DRS}. These kernels slide along the spatial and temporal dimensions of an input video sample in steps determined by the stride in each dimension~\citep{elassal:hal-03679597IJCNN}, allowing the 3D convolutional SNN to extract spatio-temporal features from the video naturally and without any pre-processing. 3D spiking convolution can be formalized as: 
\begin{equation}
v_{x,y,z,k}(t) =\sum_{n \in \mathcal{N}} W_{i(x_{n}), j(y_{n}), m(z_{n}), k_{n}, k} \times f_{s}(t-t_{n})
\end{equation}
where $v(t)$ is the potential of the neuron membrane  at time $t$, and $x$, $y$, $z$, and $k$ are the coordinates of the spike in the width, height, time, and channel dimensions, respectively. $\mathcal{N}$ is the set of input connections in the neighbourhood, $W$ is the trainable synaptic weight matrix, $i()$, $j()$, and $m()$ are functions that are used to map the location of the input spike to the corresponding location in the weight matrix, and $k$ is the index of the trainable filter. When the membrane potential $v_{x,y,z,k}(t)$ crosses the threshold potential $v_{\text{th}}(t)$, the synaptic weights and thresholds of the network are updated.
Here, we note that both the biological STDP rule as well as the threshold adaptation rule are unchanged for the 2D and 3D architectures, because these rules are independent of the dimensions of the input and of the filter. 

\section{Evaluation}
\label{section:evaluation}
This section provides the details of our experiments. First, we present the datasets, along with the implementation details and the main hyper-parameters of our network. Then, we show the results of implementing and testing our spiking two-stream neural network, where we evaluate the individual performance of the spatial and temporal streams, in addition to the performance of the resulting fused feature maps.

\subsection{Datasets and evaluation protocol}
\label{subsection:dataset}
The KTH and Weizmann datasets are early and simple datasets comparable to MNIST, but for action recognition. Although traditional computer vision approaches have already achieved high recognition rates on the these datasets~\citep{KTHGoodPErcent}, their simplicity makes them a good basic benchmark to study the performance of new models like the ones targeted in this paper. 

The KTH dataset contains 600 videos that are made up of 25 subjects performing 6 actions in 4 scenarios. The subjects 11, 12, 13, 14, 15, 16, 17 and 18 are used for training, while 19, 20, 21, 23, 24, 25, 01, 04 are used for validation and 02, 03, 05, 06, 07, 08, 09, 10 and 22 are used for testing, as indicated in the KTH protocol.

The Weizmann dataset contains 90 videos of 9 subjects performing 10 actions. The experiments on this dataset are all done using the leave-one-subject-out strategy. 

The IXMAS action recognition dataset is made up of 11 actions and 1148 sequences with different actors, cameras, and viewpoints, which brings additional complexity. Two thirds of the recordings contain objects in the scene partially occluding the actors. We also use the leave-one-subject-out strategy with this dataset. 

We also use the UCF sports dataset in order to evaluate the performance of the STDP-based SNN on a more natural dataset. This dataset contains 150 videos of 10 action. One third of the videos for each action class is used for testing and the remaining two thirds are used for training. 

To shorten the running time of experiments, we take a subset of the video frames, like in~\citep{kth3d11frames} and~\citep{3DCNNHAR}, for all datasets. We sample enough frames of a video to capture the full action. Therefore, we use $10$ frames per video, skipping three frames between each two selected frames. We also scale down the frame sizes to half of their original sizes for processing speed reasons.
%, and this step can be replaced by spatial pooling. 

We measure the classification accuracy (in \%) on the test set for all experiments. Each experiment is run three times and we report the average accuracy over the three runs. Each CSNN stream is trained independently and an SVM is placed at the end of each stream, in order to have an accurate measure of performance for the feature maps before and after fusion.

\subsection{Implementation details}
\label{subsection:impdetails} 
The meta-parameters used in this work are presented in Table~\ref{table:I}. We use a 2D on-center/off-center filter as mentioned in Section~\ref{subsection:baselinearch}. Experiments with and without this filter have been conducted, and the ones with the filter gave better results. This is because SNNs need edges to learn informative visual patterns~\citep{ImprSNNTrain}.  

The two-stream CSNN uses the spatial stream of Figure~\ref{figTSsp} for all experiments with a convolutional kernel size of $5 \times 5$. The 2D temporal streams also have a convolutional kernel size of $5 \times 5$. The 3D temporal stream has a convolutional kernel size of $5 \times 5 \times 2$. 
The value of the target timestamp $\hat{t}$ discussed in Section~\ref{subsubsection:threshadapt} is specific for each dataset, and was obtained using an exhaustive search. We use a value of $\hat{t} = 0.65$ for the KTH, IXMAS and UCF sports datasets, and a value of $\hat{t} = 0.75$ for the Weizmann dataset. The size of the output feature maps is limited to $20 \times 20 \times 2$, except for the streams that use a single frame representation of motion as a pre-processing, i.e. early fusion and motion grid, in which case the feature map size is limited to  $20 \times 20 \times 1$.

The software simulator used to simulate the convolutional SNNs tested in this work is the CSNN-simulator~\citep{ImprSNNTrain}, which is a publicly available and open-source simulator. The source code for our experiments will be released publicly.

\begin{table}
\fontsize{9pt}{9pt}\selectfont
\begin{center}
 \begin{tabular}{| c |} 
 \hline
   \textbf{STDP} \\
 \hline
 $\eta_{w}= 0.1$, $\tau_{\mathrm{STDP}}= 0.1$, $W \sim U(0, 1)$ \\
 \hline
 \textbf{Threshold Adaptation }\\
 \hline
  $\mathrm{th}_{\mathrm{min}}= 1.0 $, $\eta_{\text{th}}= 1.0$,
 \\ $\upsilon_{\text{th}}(0) \sim G(5, 1)$ \\
 \hline
 \textbf{Difference-of-Gaussian} \\
 \hline 
 $\mathrm{s}= 7.0$, $\mathrm{\sigma_{1}}= 1.0$, $\mathrm{\sigma_{2}}= 4.0$ \\
 \hline
\end{tabular}
\end{center}
\caption{The meta-parameter values used in the experiments.}
\label{table:I}
\end{table}

\subsection{Two-stream methods}
In this section, we study the effects of using different temporal configurations with two-stream methods.
Tables~\ref{table:EF},~\ref{table:OF},~\ref{table:FS},~\ref{table:MG},~and~\ref{table:3DConv} show the results of experiments where the spatial stream processes raw video frames, and the temporal streams use early fusion, optical flow, frame subtraction, motion grid, and 3D convolution respectively.

\begin{table}
\fontsize{8pt}{8pt}\selectfont
\begin{center}
\begin{tabular}{|c | c | c | c |}  

\hline
Dataset  & Raw vid (2D conv) & EF (2D conv) & Fused \\
\hline
KTH & 54.01 $\pm1.43$ & 52.16 $\pm1.57$& \textbf{55.71} $\pm0.57$ \\ [0.5ex] 
\hline
UCF sports & 35.33 $\pm1.88$ & 26.67 $\pm0.94$ & \textbf{38.00} $\pm0.00$  \\ [0.5ex]  
\hline
IXMAS & 43.46 $\pm0.34$ & 38.44  $\pm0.63$  & \textbf{48.54}  $\pm0.27$ \\ [0.5ex] 
\hline
Weizmann & \textbf{52.31} $\pm0.91$ &	51.91 $\pm0.52$& 51.91 $\pm0.52$ \\ [0.5ex] 
\hline
\end{tabular}
\end{center}
\caption{Classification rates in \% (average $\pm$ standard deviation) of the KTH, UCF sports, IXMAS, and Weizmann datasets (10 frames per video) over 3 runs. The spatial stream uses raw frames, and the temporal stream uses Early fusion (EF).}
\label{table:EF}
\end{table}

The results in Table~\ref{table:EF} show that the performance of the spatial stream with raw videos is better than that of the temporal stream, which is pre-processed with early fusion. This is because we implement temporal pooling at the end of the spatial stream, in order to join the output feature maps together. This is equivalent to a sort of late fusion, and as shown in~\citep{elassal:hal-03263914}, late fusion has a better performance than early fusion with STDP-based CSNNs. Nevertheless, fusing both streams gave superior results than any of those streams alone with all datasets except for the Weizmann dataset. With this dataset, fusing the features together did not give any added value, and thus did not result in a better classification rate. This is a consequence of the noise that is learned by the spatial stream, as discussed later in Section~\ref{section:discussion}. 

\begin{table} 
\fontsize{8pt}{8pt}\selectfont
\begin{center}
\begin{tabular}{|c | c | c | c |}  
%\hline
% \multicolumn{4}{|c|}{Raw videos (2D conv) + Optical flow (2D conv)} \\
\hline
Dataset  & Raw vid (2D conv) & OF (2D conv) & Fused \\
\hline
KTH & 54.01 $\pm1.52$ & 61.73 $\pm1.15$& \textbf{64.51} $\pm0.78$\\ [0.5ex] 
\hline
UCF sports & 35.33 $\pm1.88$ & 38.00 $\pm0.00$& \textbf{40.00} $\pm0.00$\\ [0.5ex] 
\hline
IXMAS & 43.46 $\pm0.34$ & 61.41 $\pm0.14$& \textbf{62.52} $\pm0.56$\\ [0.5ex] 
\hline
Weizmann & 52.31 $\pm0.91$ &	65.84 $\pm0.52$& \textbf{66.21} $\pm1.31$  \\ [0.5ex] 
\hline
\end{tabular}
\end{center}
\caption{Classification rates in \% (average $\pm$ standard deviation) of the KTH, UCF sports, IXMAS, and Weizmann datasets (10 frames per video) over 3 runs. The spatial stream uses raw frames, and the temporal stream uses optical flow (OF).}
\label{table:OF}
\end{table}

The results in Table~\ref{table:OF} show that the difference in performance between the temporal stream that extracts features from optical flow and the raw spatial stream can be large (more than 10 p.p.). Yet, this does not affect the fact that the temporal and spatial recognition streams are complementary, because their fusion systematically improves the results over having only one stream or the other. Therefore, with spiking models, the streams do not have to have similar performance in order for the fusion to be beneficial. % as opposed to what was reported in the case of ANNs in~\citep{Ye_2015}. %Spiking neural networks can effectively learn different feature maps from spatio-temporal input data.

\begin{table} 
\fontsize{8pt}{8pt}\selectfont
\begin{center}
\begin{tabular}{|c | c | c | c |}  
%\hline
% \multicolumn{4}{|c|}{Raw videos (2D conv) + Frame subtraction (2D conv)} \\
\hline
Dataset  & Raw vid (2D conv) & FS (2D conv) & Fused \\
\hline
KTH & 54.01 $\pm1.52$ & 64.97 $\pm0.21$& \textbf{67.75} $\pm0.21$ \\ [0.5ex] 
\hline
UCF sports & 35.33 $\pm1.88$ & 38.33 $\pm2.05$& \textbf{38.67} $\pm0.47$\\ [0.5ex] 
\hline
IXMAS & 43.46 $\pm0.34$ & 60.93 $\pm0.46$& \textbf{61.69} $\pm1.05$ \\ [0.5ex] 
\hline
Weizmann & 52.31 $\pm0.91$ &	\textbf{71.23} $\pm0.47$&  63.45 $\pm1.27$ \\ [0.5ex] 
\hline
\end{tabular}
\end{center}
\caption{Classification rates in \% (average $\pm$ standard deviation) of the KTH, UCF sports, IXMAS, and Weizmann datasets (10 frames per video) over 3 runs. The spatial stream uses raw frames, and the temporal stream uses frame subtraction (FS).}
\label{table:FS}
\end{table}

The same applies to the results displayed in Table~\ref{table:FS}, where some experiments show a significant difference in performance between the streams. The features extracted by these two streams are diverse and complementary: one of them extracts appearance information, while the other one extracts features from edges of motion, because a DoG filter is applied after frame subtraction. These results further show that these networks leverage the strengths of each stream, resulting in improved performance when fusing spatial and temporal feature maps.

\begin{table} 
\fontsize{8pt}{8pt}\selectfont
\begin{center}
\begin{tabular}{|c | c | c | c |}  
%\hline
% \multicolumn{4}{|c|}{Raw videos (2D conv) + Motion grid (2D conv)} \\
\hline
Dataset  & Raw vid (2D conv) & MG (2D conv) & Fused \\
\hline
KTH & 54.01 $\pm1.52$ & 66.82 $\pm0.21$& \textbf{67.28} $\pm0.21$\\ [0.5ex] 
\hline
UCF sports & 35.33 $\pm1.88$ & 46.00 $\pm1.63$& \textbf{50.67} $\pm2.49$\\ [0.5ex] 
\hline
IXMAS & 43.46 $\pm0.34$ &	60.50 $\pm0.12$& \textbf{63.83}  $\pm0.52$\\ [0.5ex] 
\hline
Weizmann & 52.31 $\pm0.91$ & \textbf{70.77}  $\pm0.90$& 68.52 $\pm0.80$\\ [0.5ex] 
\hline
\end{tabular}
\end{center}
\caption{Classification rates in \% (average $\pm$ standard deviation) of the KTH, UCF sports, IXMAS, and Weizmann datasets (10 frames per video) over 3 runs. The spatial stream uses raw frames, and the temporal stream uses the Motion Grid (MG) pre-processing.}
\label{table:MG}
\end{table}

Table~\ref{table:MG} shows the result of fusing the spatial stream feature maps with those extracted by using a single frame representation of motion. The same behavior as above is observed: the temporal stream that uses the motion grid pre-processing significantly outperforms the spatial one, and yet the fusion performance is still ameliorated.

\begin{table} 
\fontsize{8pt}{8pt}\selectfont
\begin{center}
\begin{tabular}{|c | c | c | c |}  
\hline
Dataset  & Raw vid (2D conv) & Raw vid (3D conv) & Fused \\
\hline
KTH & 54.01 $\pm1.52$ & \textbf{61.27} $\pm0.57$ & 54.48 $\pm1.74
$\\ [0.5ex] 
\hline
UCF sports & 35.33 $\pm1.88$ & \textbf{52.00} $\pm0.00$  & 48.67 $\pm1.88$ \\ [0.5ex] 
\hline
IXMAS & 43.46 $\pm0.34$ & 55.33 $\pm0.52$& \textbf{55.54} $\pm0.34$ \\ [0.5ex]
\hline
Weizmann & 52.31 $\pm0.91$ &	\textbf{55.41} $\pm0.59$& 55.07 $\pm0.92$ \\ [0.5ex] 
\hline
\end{tabular}
\end{center}
\caption{Classification rates in \% (average $\pm$ standard deviation) of the KTH, UCF sports, IXMAS, and Weizmann datasets (10 frames per video) over 3 runs. The spatial stream uses raw frames, and the temporal stream uses 3D Convolution.}
\label{table:3DConv}
\end{table}
Table~\ref{table:3DConv} shows the results of combining the features extracted by a 2D CSNN with those of a 3D CSNN. These results show that combining spatial features with spatio-temporal features does not systematically show notable improvement with all datasets. This is because the 3D CSNN is already able to extract the appearance information extracted by the spatial stream, in addition to temporal features. Therefore, adding the spatial features coming from the spatial stream to the spatio-temporal one results in information redundancy that does not always help ameliorate the results. However, the results do ameliorate after fusion with the IXMAS dataset. This is because convolutional neural networks are less effective when extracting spatial information from datasets with different viewpoints, like the IXMAS dataset, because very different image observations can be obtained from observing the same action from different viewpoints~\citep{POPPE2010976}. What was observed in~\citep{POPPE2010976} still applies for STDP-based learning, where having several viewpoints increases the number of patterns to learn. This decreases the feature redundancy, and thus makes the fusion beneficial.

It is interesting to check if the same applies to adding additional temporal information for some temporal redundancy. Therefore, Table~\ref{table:SP3D} shows the results of the fusion between a stream pre-processed using frame subtraction and another that uses 3D convolution. These results show that redundant temporal information does not systematically ameliorate the fusion performance. Again, the only recorded amelioration is with the IXMAS dataset due to an increased number of patterns to learn from several viewpoints. These results support the initial claim that adding redundant information regardless of its nature (spatial or temporal) to spatio-temporal information does not always result in better performance.

\begin{table} 
\fontsize{8pt}{8pt}\selectfont
\begin{center}
\begin{tabular}{|c | c | c | c |}  
%\hline
% \multicolumn{4}{|c|}{Frame subtraction (2D conv) + Raw videos (3D conv)} \\
\hline
Dataset  & FS (2D conv) &  Raw vid (3D conv) & Fused\\
\hline
 KTH & \textbf{64.97} $\pm0.21$ & 61.27 $\pm0.57$ & 64.35 $\pm0.15$\\ [0.5ex] 
\hline
UCF sports & 38.33 $\pm2.05$& \textbf{52.00} $\pm0.00$ & 43.33 $\pm2.49$\\ [0.5ex] 
\hline
IXMAS & 60.93 $\pm0.46$ &	55.33 $\pm0.52$& \textbf{62.39} $\pm1.15$ \\ [0.5ex] 
\hline
Weizmann & \textbf{71.23}  $\pm0.47$&	55.41 $\pm0.59$& 60.31 $\pm0.52$ \\ [0.5ex] 
\hline
\end{tabular}
\end{center}
\caption{Classification rates in \% (average $\pm$ standard deviation) of the KTH, UCF sports, IXMAS and Weizmann datasets (10 frames per video) over 3 runs. The spatial stream uses frames pre-processed with frame subtraction with 2D convolution, and the temporal stream uses frame subtraction with 3D convolution.}
\label{table:SP3D}
\end{table}

\section{Discussion}
\label{section:discussion}

Results show that the stream performance of the spiking networks do not have to be similar so that their fusion achieves better performance. Moreover, fusing spatial and temporal information is beneficial, even when the temporal stream records less performance than the spatial one. These results show that spiking models can effectively represent spatial as well as temporal information with STDP, and that the fusion performance is related to the nature of the fused features. This is visible when fusing spatial information obtained from a raw video with velocity information obtained from the same video pre-processed with optical flow extraction. This delivers a richness in the resulting fused spatio-temporal features, and thus ameliorates the results. On the other hand, fusing spatial features with spatio-temporal features gives no improvement in performance, as shown in Tables~\ref{table:3DConv}~and~\ref{table:SP3D}.

\begin{figure}
\centerline{\includegraphics[scale=0.35]{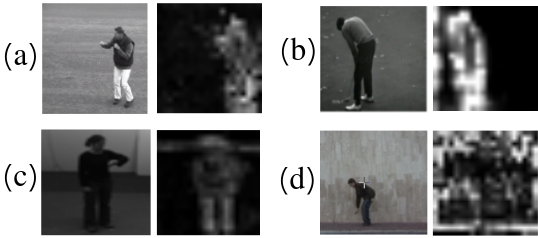}}
\caption{Raw videos and spatial stream feature maps of the (a) KTH, (b) UCF sports dataset, (c) IXMAS, and (d) Weizmann datasets.}
\label{fig:featuremaps}
\end{figure}

The results of applying the two-stream method with the Weizmann dataset in Tables~\ref{table:EF},~\ref{table:FS} and~\ref{table:MG}, show a decline in performance after fusion. Therefore, in order to understand this unsuccessful fusion, we show the extracted feature maps in Figure~\ref{fig:featuremaps}. We can see that there is a significant amount of background noise learned by the SNN with the Weizmann dataset, in comparison with the other datasets. We attempted reducing this noise by applying a Gaussian filter to both the spatial and temporal streams. The noise has slightly decreased but was not fully removed, and it still affected the fusion result. Therefore, we resort to a different method for noise reduction. We use a minimum pixel intensity as a cutoff value for the DoG filter, thus eliminating all of the pixels whose intensity is lower than this value. 

\begin{table} 
\fontsize{8pt}{8pt}\selectfont
\begin{center}
\begin{tabular}{|c | c | c | c|}  
\hline
 \multicolumn{3}{|c|}{ (a) Cutoff = 10 }\\ 
\hline
 Raw vid (2D conv) & EF (2D conv) & Fused \\
\hline
 59.32 $\pm0.00$& 65.61 $\pm0.52$& \textbf{67.35} $\pm0.00$\\ [0.5ex] 
\hline
 Raw vid (2D conv) & OF (2D conv) & Fused \\
\hline
 59.32 $\pm0.00$ & 60.43 $\pm0.00$& \textbf{62.91} $\pm0.00$\\ [0.5ex] 
 \hline
  Raw vid (2D conv) & FS (2D conv) & Fused \\
\hline
 59.32 $\pm0.00$& \textbf{59.91} $\pm0.00$& 58.80 $\pm0.00$\\ [0.5ex] 
 \hline
  Raw vid (2D conv) & MG (2D conv) & Fused \\
\hline
 59.32 $\pm0.00$& \textbf{72.17} $\pm0.40$& 59.57 $\pm0.00$\\ [0.5ex] 
 \hline
  Raw vid (2D conv) & Raw vid (3D conv) & Fused \\
\hline
 \textbf{59.32} $\pm0.00$& 55.61 $\pm1.04$& 57.58 $\pm0.52$\\ [0.5ex] 
 \hline
 \multicolumn{3}{|c|}{ (b) Cutoff = 20 }\\ 
\hline
 Raw vid (2D conv) & EF (2D conv) & Fused \\
\hline
 47.83 $\pm0.52$& 48.75 $\pm1.27$& \textbf{50.43} $\pm0.00$\\ [0.5ex] 
\hline
 Raw vid (2D conv) & OF (2D conv) & Fused \\
\hline
 47.83 $\pm0.52$& 57.21 $\pm0.52$& \textbf{59.69} $\pm0.52$\\ [0.5ex] 
 \hline
  Raw vid (2D conv) & FS (2D conv) & Fused \\
\hline
 47.83 $\pm0.52$& 48.21 $\pm0.00$ & \textbf{51.05} $\pm0.52$ \\ [0.5ex] 
 \hline
  Raw vid (2D conv) & MG (2D conv) & Fused \\
\hline
 47.83 $\pm0.52$& 30.83 $\pm3.40$& \textbf{48.72} $\pm0.00$\\ [0.5ex] 
 \hline
  Raw vid (2D conv) & Raw vid (3D conv) & Fused \\
\hline
 47.83 $\pm0.52$& \textbf{50.71} $\pm2.17$& 48.66 $\pm0.64$\\ [0.5ex] 
\hline
\end{tabular}
\end{center}
\caption{Classification rates in \% (average $\pm$ standard deviation) of the Weizmann dataset (10 frames per video) over 3 runs. (a) A minimum pixel intensity of 10, (b) A minimum pixel intensity of 20. }
\label{table:cutoff}
\end{table}

\begin{figure}
\centerline{\includegraphics[scale=0.28]{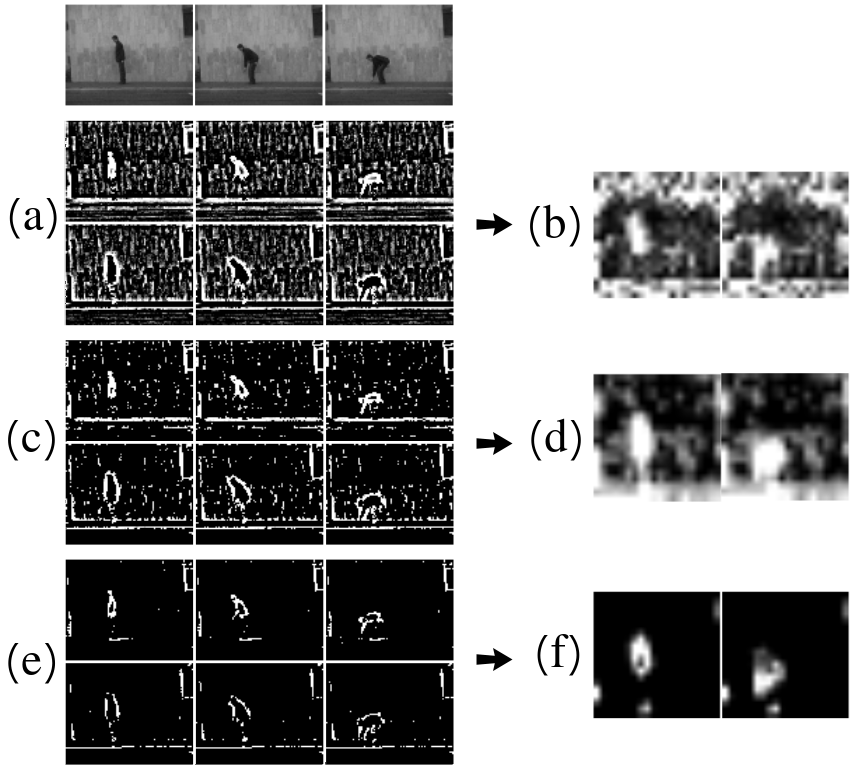}}
\caption{A video sample filtered using a DoG filter. We display both the on and off channels with (a) no minimum pixel intensity (b) its corresponding extracted feature map (c) a minimum pixel intensity of 10 (d) its corresponding feature map  (e) a minimum pixel intensity of 20 (f) its corresponding feature maps.}
\label{fig:input-cuttoff}
\end{figure}

When using a minimum pixel intensity of $10$, the classification rates of the individual streams increase as a result of noise reduction as shown in Table~\ref{table:cutoff}. This occurs with most streams, except the one for frame subtraction. The spatial noise with this stream is not significant; therefore, the loss of information from the cuttoff has a greater degrading impact on the performance than the gain of performance attributed to noise reduction. With a cutoff value of $10$ the noise is not fully removed as shown in Figure~\ref{fig:input-cuttoff} (d), and therefore the fusion is still ineffective. 
However, when increasing the minimum pixel intensity to $20$ the classification rates of the individual streams are significantly decreased as shown in Table~\ref{table:cutoff}. This is a result of discarding a lot of the information while trying to discard the noise. However, with this configuration, the fusion operation is no longer affected by the spatial noise. This successful fusion performance manifests the effects of noise on spiking two-stream methods. This indicates that there is a need for alternative noise reduction methods that improve the feature fusion without compromising the information. Moreover, the feature fusion with 3D convolution is still unsuccessful because the extracted features are still redundant in both streams.

\section{Conclusion}
\label{section:conclusion}
Video analysis is an important yet challenging task in computer vision. Current research is mostly based on deep learning methods that have high computational costs. However, spiking neural networks could be solutions that would be more energy efficient on neuromorphic hardware. In this work we focused on transposing a popular deep learning method to the spiking domain. We present spiking two-stream methods, which allows 2D CSNNs to extract spatio-temporal information from videos: we implement two-stream methods and a number of spatio-temporal and temporal streams using STDP-based CSNNs, and we gather the following conclusions:
\begin{itemize} 
  \item spiking two-stream methods can successfully extract spatio-temporal features from action recognition videos; 
\item spiking spatial and temporal streams are complementary with all of the chosen temporal stream configurations;
\item spatio-temporal streams (like 3D CSNNs) lead to information redundancy and does not give effective results with STDP-based two-stream spiking models;
\item these models are sensitive to significant spatial noise which has a negative impact on the performance.
\end{itemize}
The two-stream methods tested in this work use classical pre-processing methods that are non-spiking. Therefore future work shall focus on creating a fully spiking two-stream solution, with requires the development of spiking pre-processing methods for motion extraction and a spiking classifier. 

\section{Acknowledgments}
This work has been partially funded by IRCICA (USR 3380) under the bio-inspired project. This work has also been funded by Région Hauts-de-France.



%%       \citet{<label>} ==> Jones et al. [21]
%%       \citep{<label>} ==> [21]
%%

%  \bibliographystyle{elsarticle-num-names} 
%  \bibliography{CitationLibrary}
\bibliography{CitationLibrary}
\bibliographystyle{elsarticle-num-names}





% \end{thebibliography}
\end{document}